\documentclass[10pt,twocolumn,letterpaper]{article}

\usepackage{cvpr}
\usepackage{times}
\usepackage{epsfig}
\usepackage{graphicx}
\usepackage{amsmath}
\usepackage{amssymb}
\usepackage{booktabs}
\usepackage{enumitem}

\usepackage[pagebackref=true,breaklinks=true,letterpaper=true,colorlinks,bookmarks=false]{hyperref}

\usepackage{xcolor,colortbl}

\newcommand{\dataset}{\mbox{DiPART}}
\newcommand{\datasettwo}{\mbox{Pascal Part Matching}}

\newcommand\blfootnote[1]{%
  \begingroup
  \renewcommand\thefootnote{}\footnote{#1}%
  \addtocounter{footnote}{-1}%
  \endgroup
}

\cvprfinalcopy 

\def\cvprPaperID{2732} 

\ifcvprfinal\fi
\begin{document}

\title{Structured Set Matching Networks for One-Shot Part Labeling}

\author{Jonghyun Choi$^\star$*~~~~~Jayant Krishnamurthy$^\star$$^\dagger$~~~~~Aniruddha Kembhavi*~~~~~Ali Farhadi*$^\ddagger$
\vspace{0.4em}\\
Allen Institute for Artificial Intelligence*~~~~~~Semantic Machines$^\dagger$~~~~~~University of Washington$^\ddagger$\\
{\tt\small jonghyunc@allenai.org~~~jayant@semanticmachines.com~~~anik@allenai.org~~ali@cs.uw.edu}
}

\maketitle

\begin{abstract}
Diagrams often depict complex phenomena and serve as a good test bed for visual and textual reasoning. However, understanding diagrams using natural image understanding approaches requires large training datasets of diagrams, which are very hard to obtain. Instead, this can be addressed as a matching problem either between labeled diagrams, images or both. This problem is very challenging since the absence of significant color and texture renders local cues ambiguous and requires global reasoning. We consider the problem of one-shot part labeling: labeling multiple parts of an object in a target image given only a single source image of that category. For this set-to-set matching problem, we introduce the Structured Set Matching Network (SSMN), a structured prediction model that incorporates convolutional neural networks. The SSMN is trained using global normalization to maximize local match scores between corresponding elements and a global consistency score among all matched elements, while also enforcing a matching constraint between the two sets. The SSMN significantly outperforms several strong baselines on three label transfer scenarios: diagram-to-diagram, evaluated on a new diagram dataset of over 200 categories; image-to-image, evaluated on a dataset built on top of the Pascal Part Dataset; and image-to-diagram, evaluated on transferring labels across these datasets.

\end{abstract}
\vspace{-1em}

\blfootnote{\hspace{-2em} \noindent $^\star$ indicates equal contribution. Majority of the work has been done while JK is in AI2.}

\section{Introduction}
\label{sec:intro}

A considerable portion of visual data consists of illustrations including diagrams, maps, sketches, paintings and infographics, which afford unique challenges from a computer vision perspective. While computer vision research has largely focused on understanding natural images, there has been a recent renewal of interest in understanding visual illustrations~\cite{Kembhavi2016ADI, Ouyang2016ForgetMeNotMF, Yu2016SketchMT, Wang2015Sketchbased3S, Yu2016SketchaNetAD, sketchy2016, Zhang2011CoupledIE, Liu2015Rent3DFP}. Science and math diagrams are a particularly interesting subset of visual illustrations, because they often depict complex phenomena grounded in well defined curricula, and serve as a good test bed for visual and textual reasoning~\cite{Kembhavi2016ADI, Seo2014DiagramUI, Seo2015SolvingGP, Krishnamurthy2016SemanticPT, Ghosh2017ContextualRF}.

\begin{figure}[t]
\centering
\includegraphics[width=1\linewidth,bb=4 3 487 677]{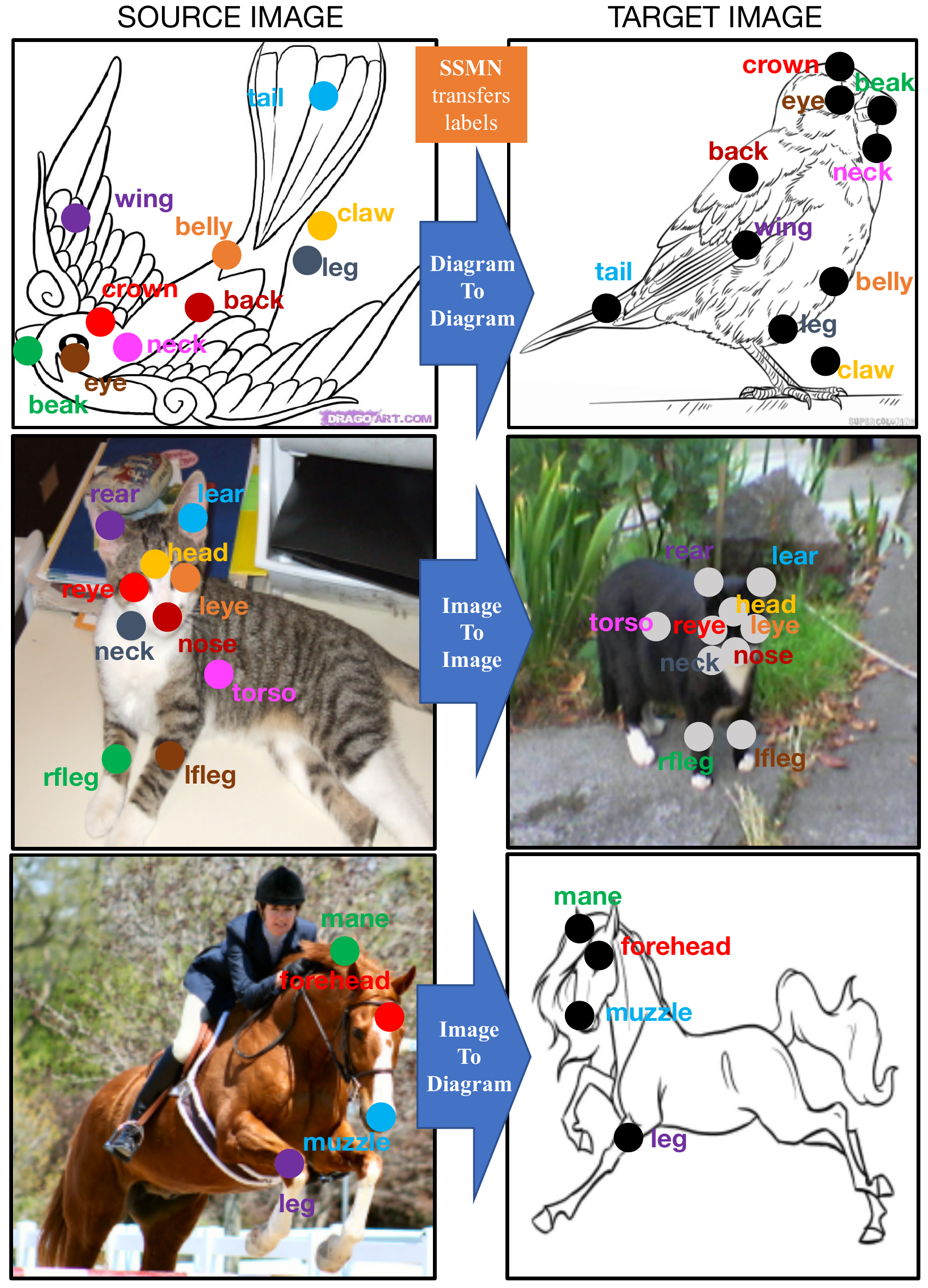}
\caption{\small Matching results by our SSMN model. Given source images annotated by points with text labels, our model transfers labels to the points in the target images. Colors indicate each label. Black (Gray in row 2) dots indicate unlabeled points. SSMN is able to correctly label the target points in spite of significant geometric transformations and appearance differences between object parts in the source and target images of categories unobserved in training.}
\vspace{-1.5em}
\label{fig:intro}
\end{figure}

Understanding diagrams using natural image understanding approaches requires training models for diagram categories, object categories, part categories, etc. which requires large training corpora that are particularly hard to obtain for diagrams.  Instead, this can be addressed by transferring labels from smaller labeled datasets of diagrams (within-domain) as well as from labeled datasets of natural images (cross-domain). Label transfer has previously shown impressive results in a within-domain natural image setting~\cite{Liu2011NonparametricSP}. It is interesting to note that young children are able to correctly identify diagrams of objects and their parts, having seen just a few diagrammatic and natural image examples in story books and textbooks. 

The task of label transfer is quite challenging, especially in diagrams. First, it requires a fine grained analysis of a diagram, but the absence of significant color or textural information renders local appearance cues inherently ambiguous. Second, overcoming these local ambiguities requires reasoning about the entire structure of the diagram, which is challenging. Finally, large datasets of object diagrams with fine grained part annotations, spanning the entire set of objects we are interested in, are expensive to acquire. Motivated by these challenges, we present the One-Shot Part Labeling task: labeling object parts in a diagram having seen only one labeled image from that category. 


One-Shot Part Labeling is the task of matching elements of two sets: the fully-labeled parts of a source image and the unlabeled parts of a target image. Although previous work has considered matching a single target to a set of sources \cite{Koch2015SiameseNN,vinyalsBLKW16}, there is little prior work on set-to-set matching, which poses additional challenges as the model must predict a one-to-one matching. For this setting, we propose the Structure Set Matching Network (SSMN), a model that leverages the matching structure to improve accuracy. Our key observation is that a matching implies a transformation between the source and target objects and not all transformations are equally likely. For example, in Figure \ref{fig:intro} (top), the matching would be highly implausible if we swapped the labels of ``wing'' and ``tail,'' as this would imply a strange deformation of the depicted bird. However, transformations such as rotations and perspective shifts are common. The SSMN \emph{learns} which transformations are likely and uses this information to improve its predictions.

The Structured Set Matching Network (SSMN) is an end-to-end learning model for matching the elements in two sets. The model combines convolutional neural networks (CNNs) into a structured prediction model. The CNNs extract local appearance features of parts from the source and target images. The structured prediction model maximises local matching scores (derived from the CNNs) between corresponding elements along with a global consistency score amongst all matched elements that represents whether the source-to-target transformation is reasonable. Crucially, the model is trained with global normalization to reduce errors from \emph{label bias} \cite{lafferty2001conditional} -- roughly, model scores for points later in a sequence of predictions matter less -- which we show is guaranteed to occur for RNNs and other locally-normalized models in set-to-set matching (Sec.\ref{sec:training}).

Off-the-shelf CNNs perform poorly on extracting features from diagrams~\cite{Kembhavi2016ADI, Yu2016SketchaNetAD}, owing to the fact that diagrams are very sparse and have little to no texture. Our key insight to overcoming this is to convert diagrams to distance transform images. The distance transform introduces \textit{meaningful} textures into the images that capture the location and orientation of nearby edges. Our experiments show that this introduced texture improves performance and enables the use of model architectures built for natural images.

We compile three datasets: (1) a new diagram dataset named Diagram Part Labeling (\dataset), which consists of 4,921 diagram images across 200 objects categories, each annotated with 10 parts. (2) a natural image part labeling dataset named Pascal Part Matching (PPM) built on top of the popular Pascal Part dataset~\cite{chenMLFUY14}. (3) a combination of the above two datasets (Cross-DiPART-PPM) to evaluate the task of cross-domain label transfer. The SSMN significantly outperforms several strong baselines on all three datasets. 

In summary, our contributions include: (a) presenting the task of One-Shot Diagram Part Labeling (b) proposing Structured Set Matching Networks, an end-to-end combination of CNNs and structured prediction for matching elements in two sets (c) proposing converting diagrams into distance transforms, prior to passing them through a CNN (d) presenting a new diagram dataset \dataset\ towards the task of one-shot part labeling (e) obtaining state-of-the-art results on 3 challenging setups: diagram-to-diagram, image-to-image and image-to-diagram.

\section{Related Work}
\label{sec:background}

\noindent \textbf{One-Shot Learning.}
Early work on one-shot learning includes Fei-Fei \etal~\cite{FeiFei2003ABA, FeiFei2006OneshotLO} who showed that one can take advantage of knowledge coming from previously learned categories, regardless of how different these categories might be. Koch \etal~\cite{Koch2015SiameseNN} proposed using a Siamese network for one-shot learning and demonstrated their model on the Omniglot dataset for character recognition. More recently Vinyals \etal~\cite{vinyalsBLKW16} proposed a matching network for one-shot learning, which incorporates additional context into the representations of each element and the similarity function using LSTMs. The SSMN model builds on matching networks by incorporating a global consistency model that improves accuracy in the set-to-set case.

\begin{figure*}[t]
\centering
\includegraphics[width=0.96\linewidth]{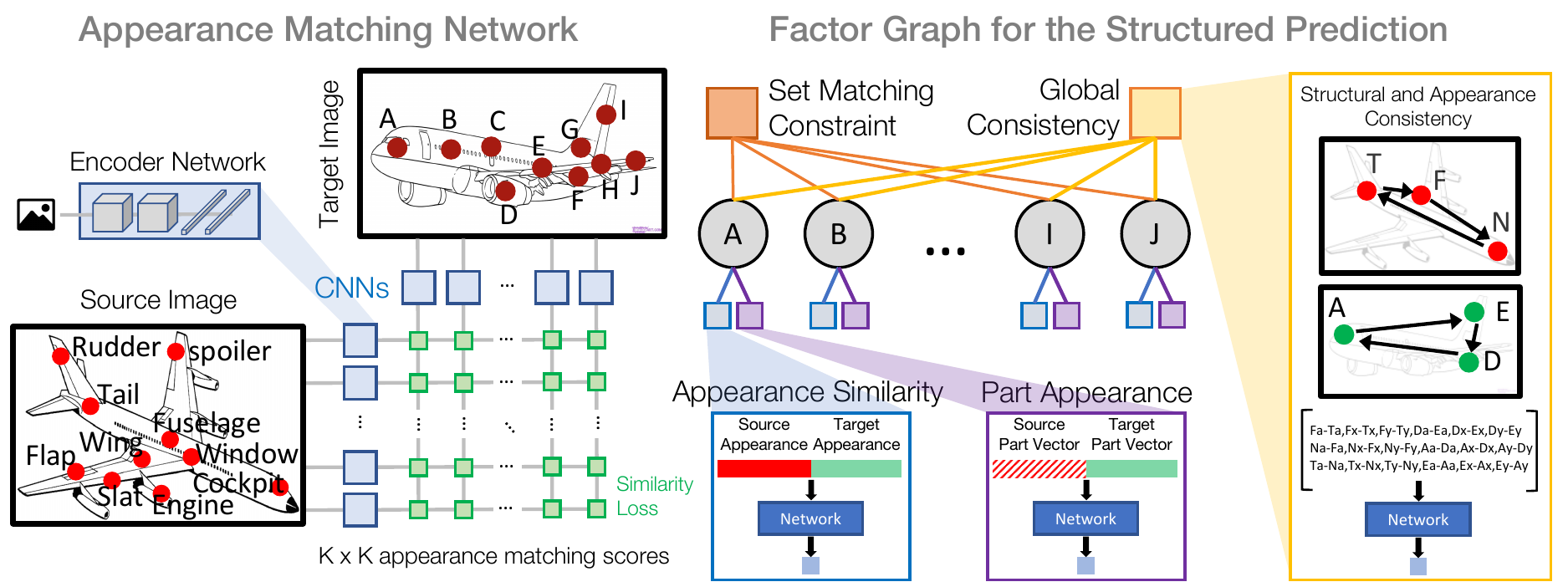}
\caption{\small Overview of the Structured Set Matching Network (SSMN) model.}
\vspace{-1em}
\label{fig:model}
\end{figure*}

\noindent \textbf{Visual Illustrations.}
There is a large body of work in sketch based image retrieval (SBIR)~\cite{Yu2016SketchMT, sketchy2016, Wang2015Sketchbased3S, Zhang2011CoupledIE, Eitz2012SketchbasedSR}. SBIR has several applications including online product searches~\cite{Yu2016SketchMT}. The key challenge in SBIR is embedding sketches and natural images into a common space, and is often solved with variants of Siamese networks. In SSMN, each pair of source and target encoders with the corresponding similarity network (Section~\ref{sec:appearance_similarity}) can be thought of as a Siamese network. There also has been work in sketch classification~\cite{eitzHA12}. More recently~\cite{Yu2016SketchaNetAD} proposed a CNN architecture to produce state-of-the-art results on this set. They noted that off-the-shelf CNN architectures do not work well for sketches, and instead proposed a few modifications. Our analysis shows that converting diagrams to distance transform images allows us to use architectures resembling ones designed for natural images. Work in understanding diagrams for question answering includes domains of science ~\cite{Kembhavi2016ADI, Krishnamurthy2016SemanticPT}, geometry~\cite{Seo2014DiagramUI, Seo2015SolvingGP} and reasoning~\cite{Ghosh2017ContextualRF}. Abstract scenes have also been analyzed to learn semantics~\cite{Zitnick2013BringingSI} and common sense~\cite{Vedantam2015LearningCS}.

\noindent \textbf{Part Recognition.}
There is a large body of work in detecting parts of objects as a step towards detecting the entire object including~\cite{chenMLFUY14, felzenszwalb2010object, Zhu2010LatentHS, Shrivastava2013BuildingPO, Azizpour2012ObjectDU, Sun2011ArticulatedPM} to name a few. In contrast to these efforts (which learn part classifiers from many training images), we focus on one-shot labeling.

\noindent \textbf{Learning Correspondences and Similarity Metrics.}
Labeling parts from a source image can be translated into a correspondence problem, which have received a lot of attention over the years. Recently, deep learning models have been employed for finding dense correspondences~\cite{Choy2016UniversalCN, ham2016, Zbontar2015ComputingTS, hanRHWCSP17} and patch correspondences~\cite{Zagoruyko2015LearningTC, Jaderberg2015SpatialTN}. The SSMN differs from the majority of them due to its ability to jointly reason over the set of elements in the source and target. There has also been a fair amount of work on learning a metric for similarity~\cite{bromleyBB93, chopraHadsellLeCun05, songXJS16}. The appearance similarity factor in the SSMN model builds on past work in this area. Recently, Han \etal~\cite{hanRHWCSP17} have proposed incorporating geometric plausibility into a model for semantic correspondence, a notion that is also strongly leveraged by the SSMN.

\noindent\textbf{Global Normalization with Neural Networks.} Most work on structured prediction with neural networks uses locally-normalized models, \eg, for caption generation \cite{karpathy2015}. Such models are less expressive than globally-normalized models (\eg, CRFs) \cite{Andor2016} and suffer from \emph{label bias} \cite{lafferty2001conditional}, which, as we show in Sec~\ref{sec:training}, is a significant problem in set-to-set matching. A few recent works have explored global normalization with neural networks for pose estimation \cite{tompson2014joint} and semantic image segmentation \cite{schwing2015fully,zheng2015}. Models that permit inference via a dynamic program, such as linear chain CRFs, can be trained with log-likelihood by implementing the inference algorithm (which is just sums and products) as part of the neural network's computation graph, then performing backpropagation \cite{Gormley2015,Eisner2016,Yatskar2016,do2010neural,peng2009conditional}. Some work has also considered using approximate inference during training \cite{chen2015learning,tompson2014joint,wang2016proximal}. Search-based learning objectives, such as early-update Perceptron \cite{Collins2004} and LaSO \cite{Daume2005}, are other training schemes for globally-normalized models that have an advantage over log-likelihood: they do not require the computation of marginals during training. This approach has recently been applied to syntactic parsing \cite{Andor2016} and machine translation \cite{Wiseman2016SequencetoSequenceLA}, and we also use it to train SSMN.


\section{Structured Set Matching Network}
\label{sec:model}

The structured set matching network (SSMN) is a model for matching elements in two sets that aims to maximise local match scores between corresponding elements and a global consistency score amongst all matched elements, while also enforcing a matching constraint between the two sets. We describe SSMN in the context of the problem of one-shot part labeling, though the model is applicable to any instance of set-to-set matching.

The one-shot part labeling problem is to label the parts of an object having seen only one example image from that category. We formulate this problem as a label transfer problem from a source to a target image. Both images are labeled with $K$ parts, each of which is a single point marked within the part as shown in Figure~\ref{fig:model}. Each part of the source image is further labeled with its name, \emph{e.g.}, ``tail.'' The model's output is an assignment of part names to the points marked in the target image, each of which much be uniquely drawn from the source image.

There are several modeling challenges in the one-shot part labeling task. First, the model must generalize to images of unseen categories, with parts that were never encountered during training. Thus, the model cannot simply learn a classifier for each part name. Second, spatially close part locations and the absence of any color or textual information in diagrams renders local appearance cues inherently ambiguous. Thus, part labeling cannot be decomposed into independent labeling decisions for each target part without losing valuable information. Third, pose variations between pairs of images renders absolute positions ambiguous. To overcome the ambiguities, the model must jointly reason about the relative positions of all the matched parts to estimate whether the pose variation is globally consistent. Finally, the model must enforce a 1:1 matching.

The SSMN addresses the above challenges by using a convolutional neural network to extract local appearance cues about the parts from the source and target images, and using a structured model to reason about the entire joint assignment of target parts to source parts without making per-part independence assumptions. It is a non-probabilistic model; thus, it is similar to a conditional random field \cite{lafferty2001conditional} with neural network factors, except its scores are not probabilities. Figure~\ref{fig:model} shows an overview of the SSMN applied to the problem of one shot part labeling. The factor graph representation shows the four factors is the SSMN:

\begin{enumerate}[noitemsep,nolistsep]
\item Appearance similarity $(f_a)$ -- Captures the local appearance similarity between a part in the source image and a part in the target image. (Sec.~\ref{sec:appearance_similarity})
\item Part appearance $(f_p)$ -- Captures the appearance similarity between a part in the target image and a name assigned to it. In the one-shot setting, this is only valuable for parts seen a priori amongst object categories in the training data. (Sec.~\ref{sec:part_appearance})
\item Global consistency $(f_{gc})$ -- Scores whether the relationships between target parts are globally consistent with those of the matched source parts, \ie. whether the source-to-target transformation is reasonable (Sec.~\ref{sec:global_consistency})
\item Matching constraint $(f_m)$ -- Enforces that target labels are matched to unique source labels.
\end{enumerate}
The first three of these factors represent neural networks whose parameters are learned during training. The fourth factor is a hard constraint. Let $m$ denote a matching where $m(i) = j$ if target part $i$ is matched to source $j$. SSMN assigns a score to $m$ using these factors as:
\begin{equation}
\resizebox{7.5cm}{!}{$
    f(m) = f_{gc}(m) + f_m(m) + \sum_i f_a(m(i), i) + f_p(m(i), i).
$}
\label{eq:ssmn_model}
\end{equation}

\subsection{Appearance Similarity}
\label{sec:appearance_similarity}

The appearance of each object part is encoded by an \emph{encoder network}, a CNN whose architecture is akin to the early layers of VGG16~\cite{simonyanZ14}. The input to the CNN is an image patch extracted around the annotated part and resized to a canonical size. The output of the CNN is an embedding of the local appearance cues for the corresponding object part. The $2K$ object parts (from both the source and target images) are each fed to $2K$ copies of the encoder with shared weights, producing $2K$ appearance embeddings. The model creates contextualized versions of these embeddings by running a \emph{context network}, a bidirectional LSTM, over the source embeddings and, as an independent sequence, the target embeddings. Note that the source and target points are given as sets, so we shuffle them arbitrarily before running the LSTMs. The similarity score between a source and target point is the dot product of their contextualized embeddings. This produces $K^2$ appearance similarity scores ($f_a$) as depicted by green boxes in Fig~\ref{fig:model}.

Due to the sparse nature of diagrams and the absence of much color and texture information, CNN models pre-trained on natural image datasets perform very poorly as encoder networks. Furthermore, off the shelf CNN architectures also perform poorly on diagrams, even when no pre-training is used~\cite{Yu2016SketchaNetAD}, and require custom modifications such as larger filter sizes. Our key insight to overcoming this problem is to convert diagrams to distance transform images, which introduces meaningful textures into the images that capture the location and orientation of nearby edges. This noticeably improves performance for diagrams, whilst using the CNN architectures designed for natural images.  

\subsection{Part Appearance}
\label{sec:part_appearance}
In the one-shot setting, at test time, the model observes a single fully labeled image from an object category, that it has not seen before. However, some common part names are likely to recur. For example, if various animal categories appear across training, validation and test categories, parts such as ``leg'' will recur. Thus, a model can benefit from learning typical appearances of these common parts across all types of images. The part appearance factor enables the model to learn this kind of information.

Let $p_i$ be a parameter vector for the $i^\text{th}$ source part's name, and $t_j$ be the output of the \emph{encoder network} for the $j^\text{th}$ target part (Sec.~\ref{sec:appearance_similarity}). The part appearance model assigns a match score $f_p(i, j)$ between source $i$ and target $j$: $f_p(i, j) = w_2^T ~\mathtt{relu}(W_1 [p_i ~ t_j]^T + b).$
Along with the layer parameters, $p_i$ is also learned at training time. The model has a separate parameter vector $p_i$ for each part name that appears at least twice in the training data; all other parts are mapped to a special ``unknown'' parameter vector.

\subsection{Global Consistency}
\label{sec:global_consistency}
In addition to local appearance, consistency of the relations between matched source and target parts provides a valuable signal for part set matching. However, these relations may be transformed in an unknown but systematic way when moving from the source to the target. For example, if the target is left-right flipped relative to the source, all parts to the left of part $x$ in the source should be on the right of $x$ in the target. Alternatively, the target may be drawn in a different style that affects the appearance of each part. Given a matching, the global consistency factor learns whether the implied source-to-target transformation is likely.

We factor the global consistency ($f_{gc}$) into the sum of two terms: structural consistency ($f_{sc}$) for pose variations and appearance consistency ($f_{ac}$) for style variations. Both terms are neural networks that score entire matchings $m$ using the same architecture, but different inputs and parameters. The score for a matching is computed from a set of \emph{relation vectors} $\Delta(m)_{ij}$ for each part pair $i,j$ in the matching $m$, then applying fully connected layers and sum-pooling: 
\vspace{-.5em}
\begin{equation}
\begin{split}
h_{ij}(m) = \mathtt{relu}(&W_2 ~ \mathtt{relu}(W_1 \Delta(m)_{ij} + b_1) + b_2),\\
f_*(m) &= \sum_i^{|m|} \sum_j^{|m|} w^T h_{ij}(m),
\end{split}
\label{eq:structural_consistency_2}
\end{equation}
where $*$ could be $sc$ or $ac$.
For structural consistency ($f_{sc}$), $\Delta(m)_{ij}$ encode the relative positions of pairs of matched parts. Recall that $m(i)$ denotes the source part matched to target part $i$. Let $loc^s_{m(i)}$ and $loc^t_i$ denote the x/y positions of source part $m(i)$ and target part $i$ respectively. The relative positions of a pair of parts $i,j$ are then encoded as a 4-dim vector, $\Delta(m)_{ij} = [ loc^s_{m(j)} - loc^s_{m(i)}, loc^t_j - loc^t_i ]$. For appearance consistency ($f_{ac}$), the $\Delta$ vectors replaced by $[ app^s_{m(j)} - app^s_{m(i)}, app^t_j - app^t_i ]$, where $app^s_{m(i)}$ and $app^t_i$ represent the appearance embeddings output by the encoder network in Sec.~\ref{sec:appearance_similarity}.


\section{Training and Inference}
\label{sec:training}
\noindent\textbf{Training.}
We train the SSMN by optimizing a structured loss on the set of part-matched images using stochastic gradient descent (SGD). Each iteration of SGD evaluates the model on a single pair of images to compute a per-example loss. Gradients are then backpropagated through the neural networks that define the model's factors.

Crucially, we train SSMN with global normalization. We found locally-normalized models performed poorly on set-to-set matching because they progressively begin to ignore scoring information as the sequence continues. A locally-normalized model, such as an RNN, would order the target points and then learn a probability distribution $P(m(i) = j | m(i-1), ..., m(1))$. After each prediction, the space of possible source points for the remaining points decreases by 1 in order to guarantee a matching. This process is problematic: note the probability for the final point is always 1, as there is only 1 source point remaining to choose from. Thus, even if the model is confident that the final pair does not match based on a pairwise similarity score, that information will be \emph{ignored entirely} in its probabilities. This problem is an instance of \emph{label bias} \cite{lafferty2001conditional}, known to reduce the accuracy of locally-normalized models. This observation is also consistent with that of Vinyals \etal~\cite{Vinyals2015OrderMS, vinyalsBLKW16}, who observed that treating unordered sets as ordered sequences enables the use of RNN models, which provide improvements to matching performance; however the ordering of elements passed to the RNNs matters.

Our training uses Learning as Search Optimization (LaSO) framework~\cite{Daume2005}, an objective function that is well-suited to training globally-normalized models with intractable exact inference. These models often rely on beam search to perform approximate inference, as does SSMN. During training, the LaSO objective penalizes the model each time the correct labeling falls off the beam, thereby training the model parameters to work well with the beam search. Also, unlike other objectives for globally-normalized models (\eg, log-likelihood of the matching), LaSO's gradient can be calculated without computing the marginal distribution over matchings or the highest-scoring matching. This is important as, in SSMN, both quantities are intractable to compute exactly due to the global consistency factor.

The LaSO objective function for a single training example is as follows. Each training example inputs to the model a pair of annotated images, and a label $m^*$ that represents the correct part matching for the pair. The LaSO objective is defined in terms of the intermediate results of a beam search with beam size $B$ in the model given the input. Let $\hat{m}^i_{t}$ denote the $i^\text{th}$ highest-scoring matching on the beam after the $t^\text{th}$ search step. Let $m^*_{t}$ denote the correct partial matching after $t$ time steps. The LaSO objective function encourages the score of $m^*_{t}$ to be higher than that of the lowest-scoring element on the beam at each time step of the search:
\begin{equation}
\mathcal{L}(f) = \sum_{t=1}^T \max(0, \Delta(m^*_t, \hat{m}^i_t) + f(\hat{m}^B_t) - f(m^*_t)).
\end{equation}

\begin{figure*}[ht]
    \centering
    \includegraphics[width=0.13\linewidth]{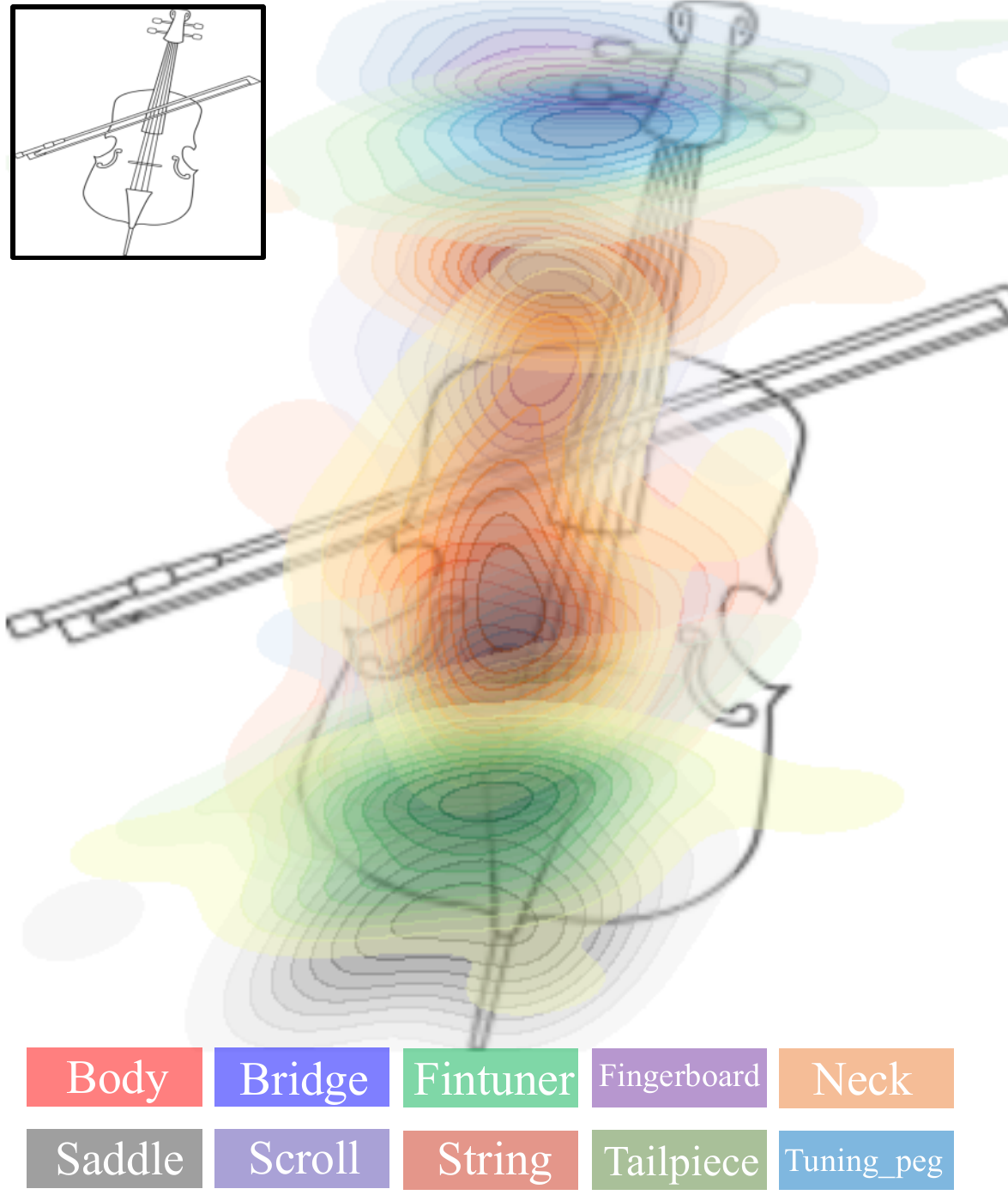}
    \includegraphics[width=0.13\linewidth]{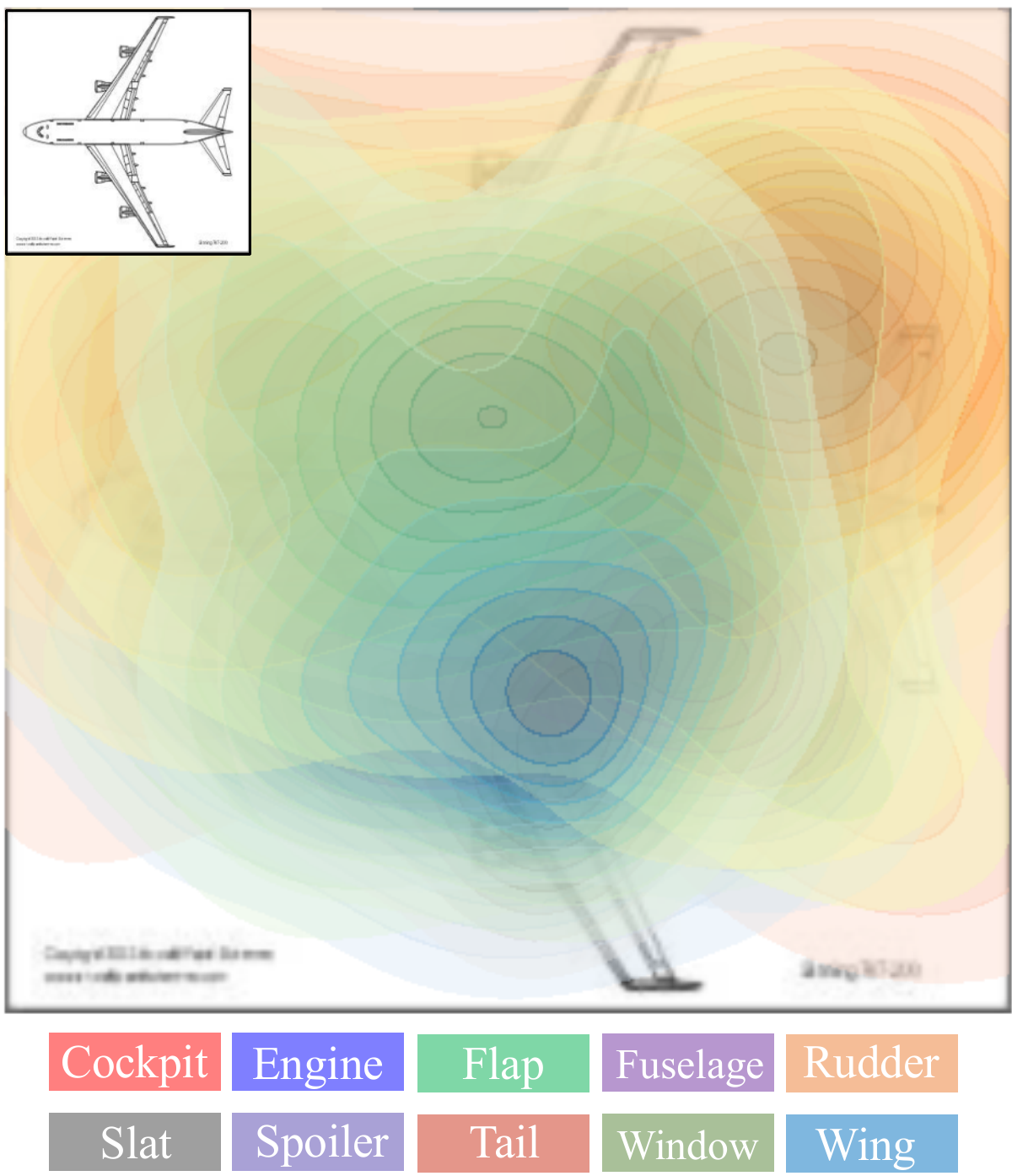}
    \hspace{.3em}
    \includegraphics[width=0.16\linewidth]{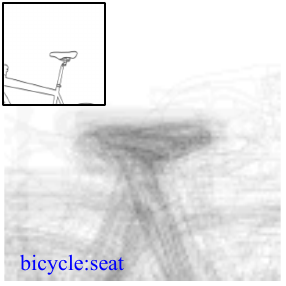}
    \includegraphics[width=0.16\linewidth]{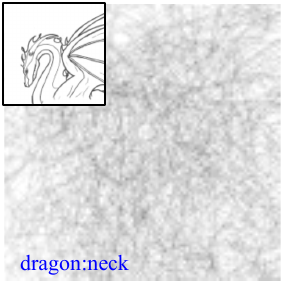}
    \hspace{.3em}
    \includegraphics[width=0.13\linewidth]{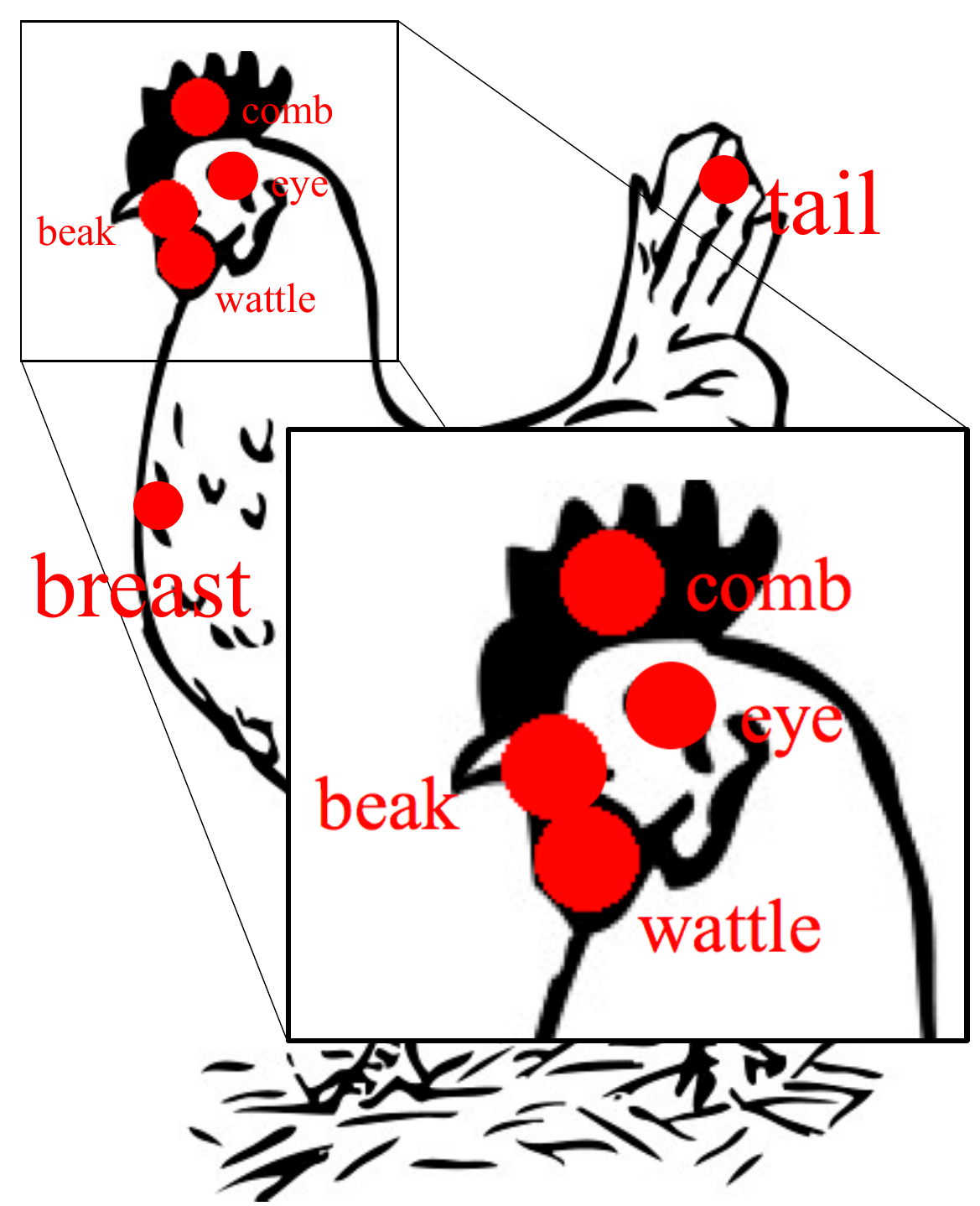}
    \includegraphics[width=0.23\linewidth]{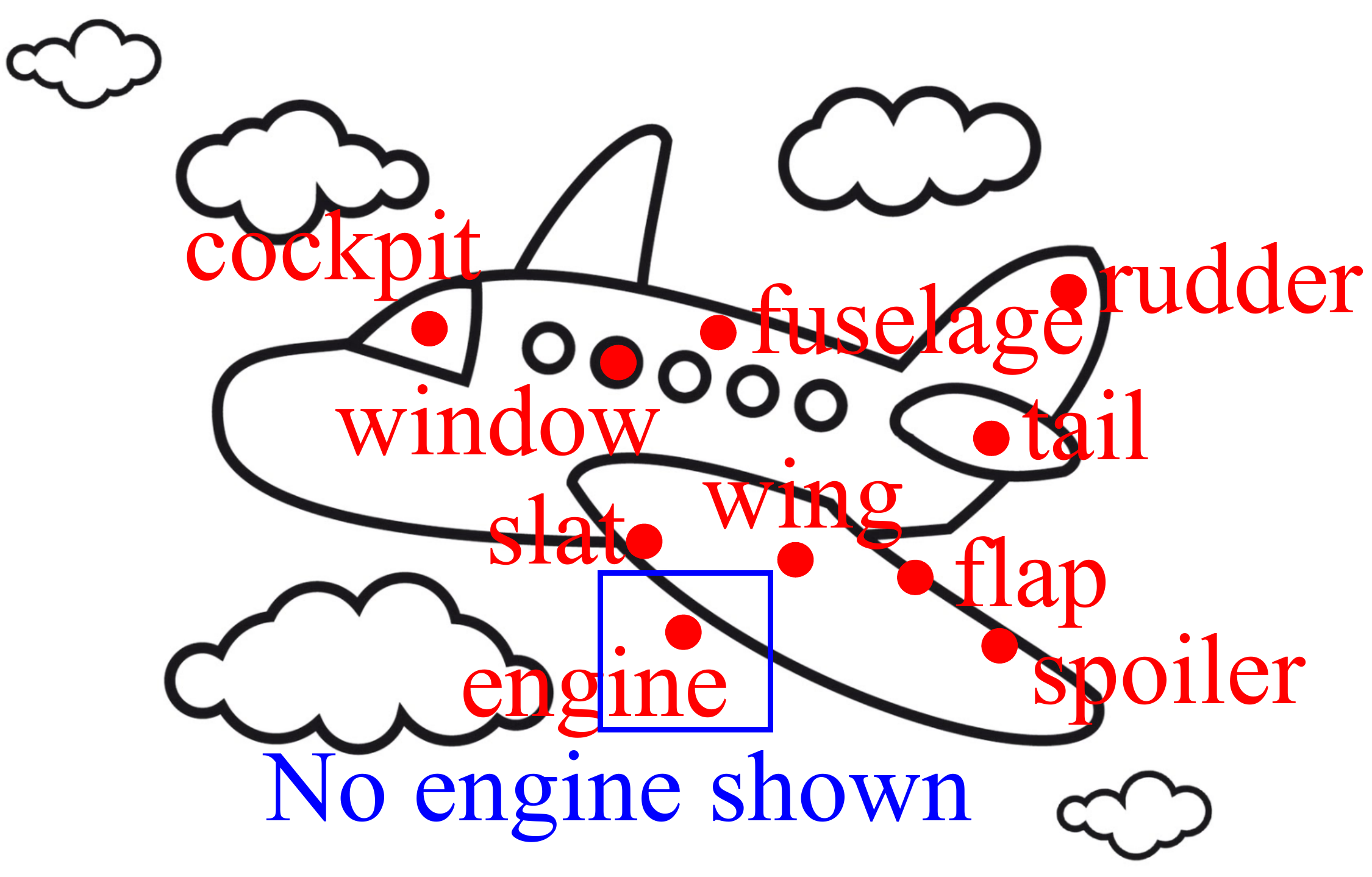}
    \resizebox{18cm}{!}{
	\begin{tabular}{ccc}
        {\small \hspace{-2em}\textbf{(a) Spatial distribution of part locations.}}&
        {\small \hspace{-2em}\textbf{(b) Average images of a part patch of an object category.}}&
        {\small \textbf{(c) Annotation examples.}}\\
        {\footnotesize (Left) Coherently (Right) Incoherently distributed parts}&
        {\footnotesize \hspace{-2em} (Left) Scale and shape coherence (Right) Incoherence}&
        {\footnotesize (Left) Spatially close parts (Right) Occluded by other parts or missing.}
    \end{tabular}
    }
    \vspace{.1em}
    \caption{Challenges in the \dataset\ dataset. Local position and appearance cues are often insufficient to provide good matching.}
    \vspace{-1em}
    \label{fig:difficult_dataset}
\end{figure*}

This loss function is a margin-based objective, similar to that of a structured SVM~\cite{Tsochantaridis2004SupportVM} or max-margin Markov network~\cite{Taskar2003MaxMarginMN}. 
The loss is 0 whenever the score of the correct partial matching $f(m^*_t)$ is greater than that of the lowest-scoring beam element $f(\hat{m}^B_t)$ by the margin $\Delta(m^*_t, \hat{m}^i_t)$, and nonzero otherwise. We set $\Delta(m^*_t, \hat{m}^i_t)$ to be the number of incorrectly matched points in $\hat{m}^i_t$ (We have omitted the dependence of $f$ on the input and model parameters for brevity). At the last time step, $B$ is set to $1$ to encourage the correct matching to have the highest score. If at any point during the search the correct partial matching falls off the beam, the search is restarted by clearing the search queue and enqueuing only the correct partial matching.

Calculating the gradient of the neural network parameters with respect to this loss function has two steps. The first step is the forward computation, which runs beam search inference in the SSMN on the input and the corresponding forward passes of its constituent neural networks. After each step of the beam search, the gradient computation checks for a margin violation. If a margin violation is found, it is recorded and the search is restarted from the correct partial matching.
If not, the beam search continues normally. The output of the forward computation is a collection of margin violations and a value for the loss function. The second step is the backward computation, which backpropagates the loss through neural networks to compute the gradient. The loss $\mathcal{L}$ is a sum of terms of the form $f(m)$, and $f(m)$ is a sum of scores output by $f$'s constituent neural networks (Equation \ref{eq:ssmn_model}). Thus, the gradient $\frac{\partial \mathcal{L}}{\partial f}$ is simply a weighted sum of the gradients of the constituent neural networks, each of which can be calculated using standard backpropagation. The inputs with respect to which the gradients are calculated, as well as each gradient's weight in the sum, depend on the the particular margin violations encountered in the forward computation.
We refer the reader to~\cite{Wiseman2016SequencetoSequenceLA} for a detailed description of the gradient computation for LaSO in neural sequence-to-sequence modeling.\footnote{
We implement SSMN using Probabilistic Neural Programs (PNP) \cite{Murray2016}, a library for structured prediction with neural networks that provides a generic implementation of LaSO. 
}

\vspace{.5em}
\noindent\textbf{Encoder Network Initialization.}
The encoder networks (Section~\ref{sec:appearance_similarity}) are pre-trained by optimizing a surrogate objective. A bank of CNNs encode image patches of parts and a bank of similarity networks compute the similarities between the appearance encodings (represented as a $K \times K$ matrix in Figure~\ref{fig:model}, where $K$ is number of parts). Each row and each column of this matrix undergo a softmax operation followed by a $K$ category cross-entropy objective. The surrogate objective is the sum of these $2K$ cross-entropy objectives. This surrogate objective encodes local appearances, but is faster to train than the SSMN objective, and is hence suitable for pre-training the appearance encoder networks. We refer to this surrogate objective as the appearance matching network (AMN) objective.

\vspace{.5em}
\noindent\textbf{Inference.}
Exact inference in SSMN is intractable due to the global consistency factor, which defines a global score for the entire matching.\footnote{Without global consistency, exact inference in SSMN can be performed with the Hungarian algorithm for maximum-weighted matching.} Thus, exact inference would require enumerating and scoring all $K!$ permutations of $K$ parts. Instead, we use approximate inference via beam search. As outlined above, SSMN is trained to ensure that the beam search is a good approximate inference strategy. The beam search starts by ordering target parts arbitrarily. The search maintains a queue of partial matchings, which at time step $i - 1$ consists of $B$ partial matchings between the first $i-1$ target parts and the source parts. The $i^\text{th}$ search step generates several new matchings for each partial matching on the queue by matching the $i^\text{th}$ target part with each unmatched source part. The search computes a score for each expanded matching and enqueues it for the next step. The search queue is then pruned to the $B$ highest-scoring partial matchings. This is repeated until each target part has been assigned to a source part label. The global consistency factor is used to score partial matchings by generating the \emph{relation vectors} (in Eq.\ref{eq:structural_consistency_2}) for the points matched thus far.

\section{Datasets}
\label{sec:dataset}
\noindent\textbf{\dataset\ Dataset.}
We present the Diagram Part Labeling (\dataset) dataset, consisting of 4,921 images across 200 object categories. Categories span rigid objects (\eg, cars) and non-rigid objects (\eg, animals). Images are obtained from Google Image Search and parts are labelled by annotators. \dataset\ is split into train, val and test sets, with no categories overlapped. Since each pair of images within a category can be chosen as a data point, the number of data points is large (101,670 train, 21,262 val, and 20,110 test).



\dataset\ is challenging for several reasons. First, the absence of color and dense texture cues in diagrams renders local appearance cues ambiguous. Second, having access to only point supervision~\cite{bearmanRFF16} at training time is challenging compared to having detailed segmentation annotations for parts as in previous natural image datasets (\eg, Pascal Part~\cite{chenMLFUY14}). Third, parts for several categories are located very close by, requiring very fine grained analysis of the texture-sparse diagrams (Fig.~\ref{fig:difficult_dataset}-(c)).
Fourth, the appearances and locations of parts are generally not coherent across samples within a category. Finally, the one-shot setting renders this even more challenging.




\vspace{.5em}
\noindent\textbf{\datasettwo\ (PPM) Dataset.}
To evaluate SSMN on labeling parts in natural images, we use images from the Pascal Part dataset~\cite{chenMLFUY14} with more than 10 parts and convert part segments to point annotations using the centers of mass. We called it Pascal Part Matching (PPM), which consists of 74,660 train and 18,120 test pairs in 8 categories with 10 parts.

\vspace{.5em}
\noindent\textbf{Cross-DiPART-PPM Dataset.}
For cross domain matching experiments, we find all overlapping categories and part names between \dataset\ and \datasettwo\ to make Cross-DiPART-PPM. It consists of 5 categories with 4 parts and 22,969 image-to-diagram pairs (18,489 train and 4,480 test).

\vspace{.5em}
\noindent More details about the datasets including download links as well as more results can be found in the \href{https://allenai.github.io/one-shot-part-labeling/}{project page}.

\section{Experiments}
\label{sec:exp}

\begin{figure*}[t]
    \includegraphics[width=0.33\linewidth]{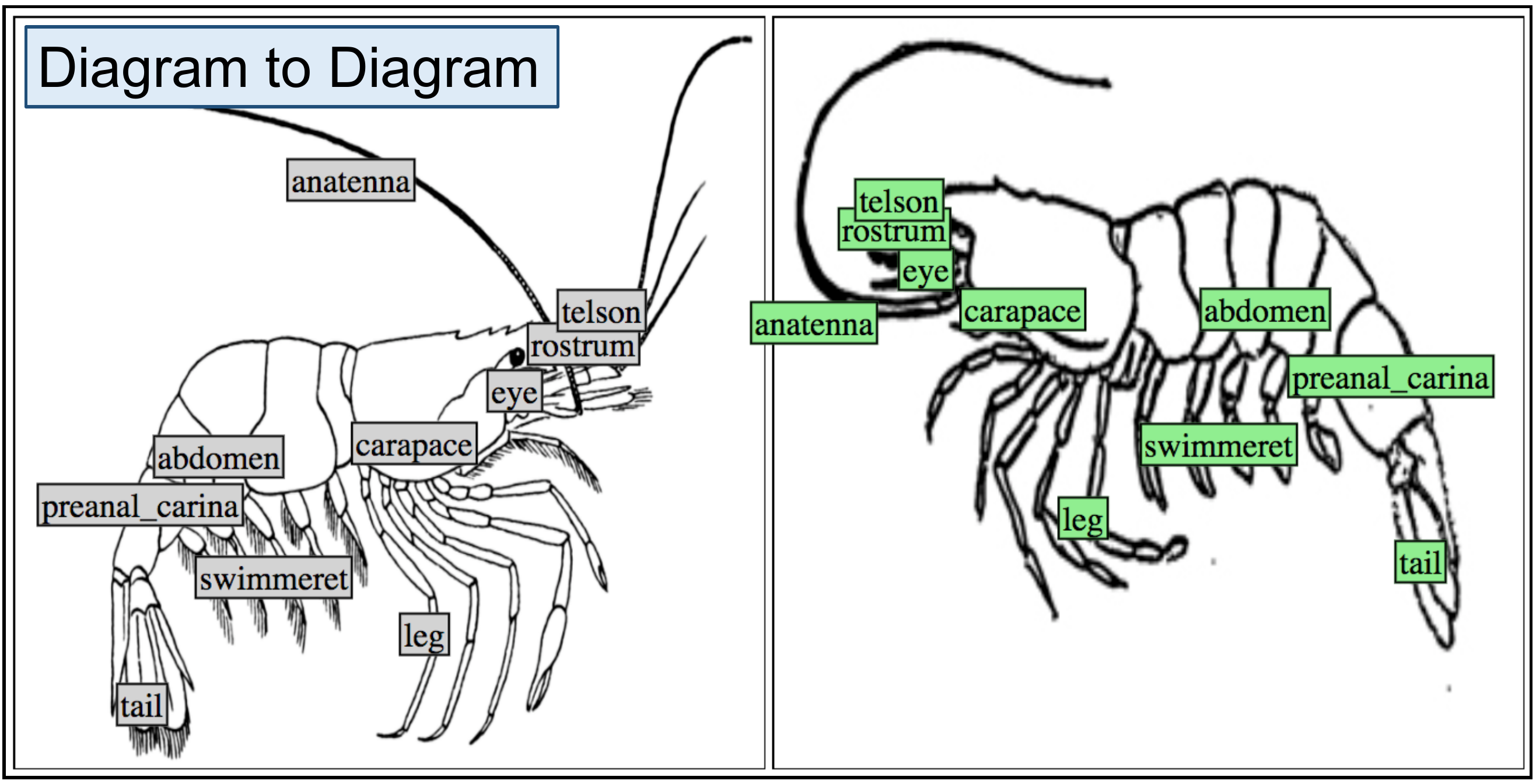}   
    \includegraphics[width=0.33\linewidth]{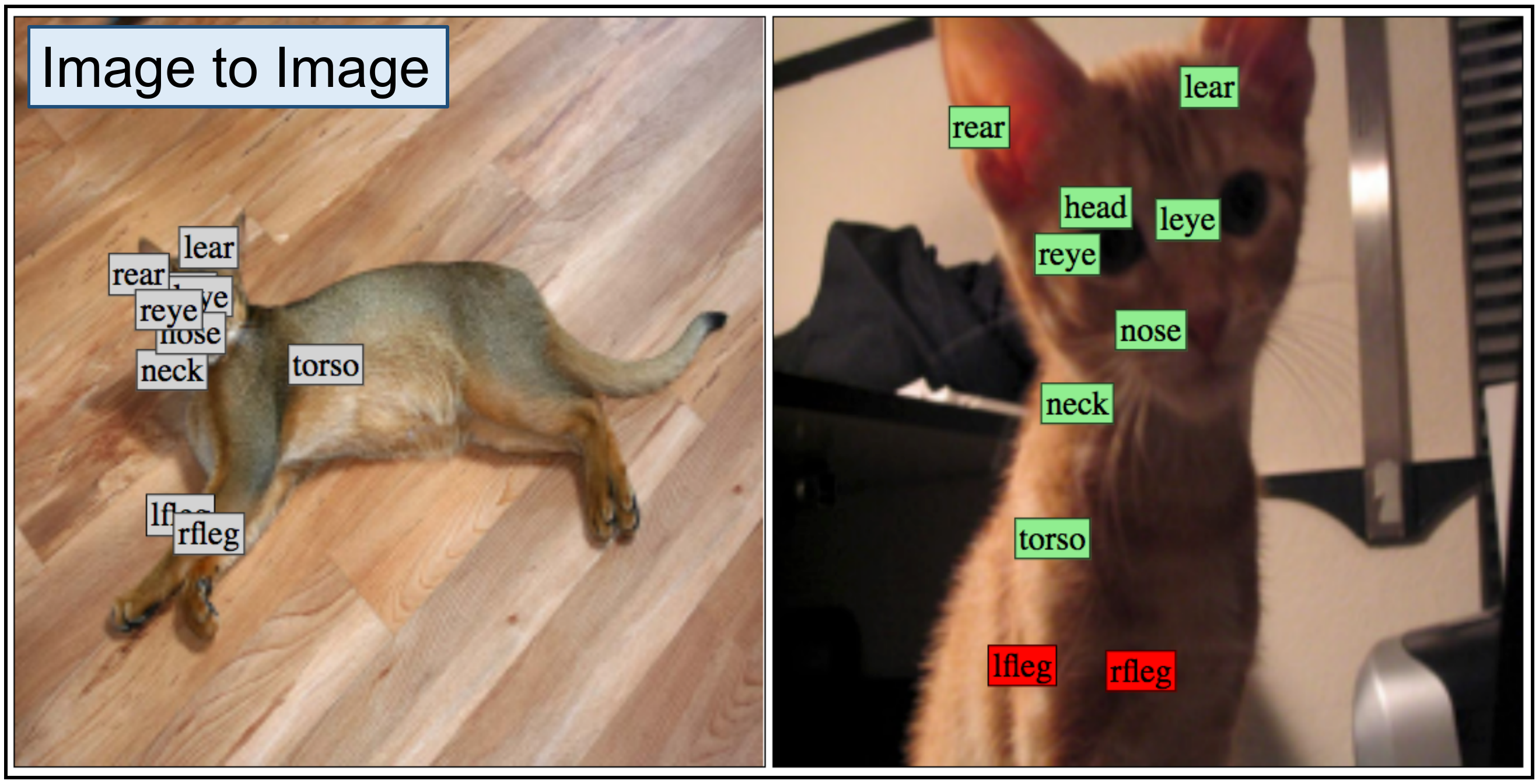}
    \includegraphics[width=0.33\linewidth]{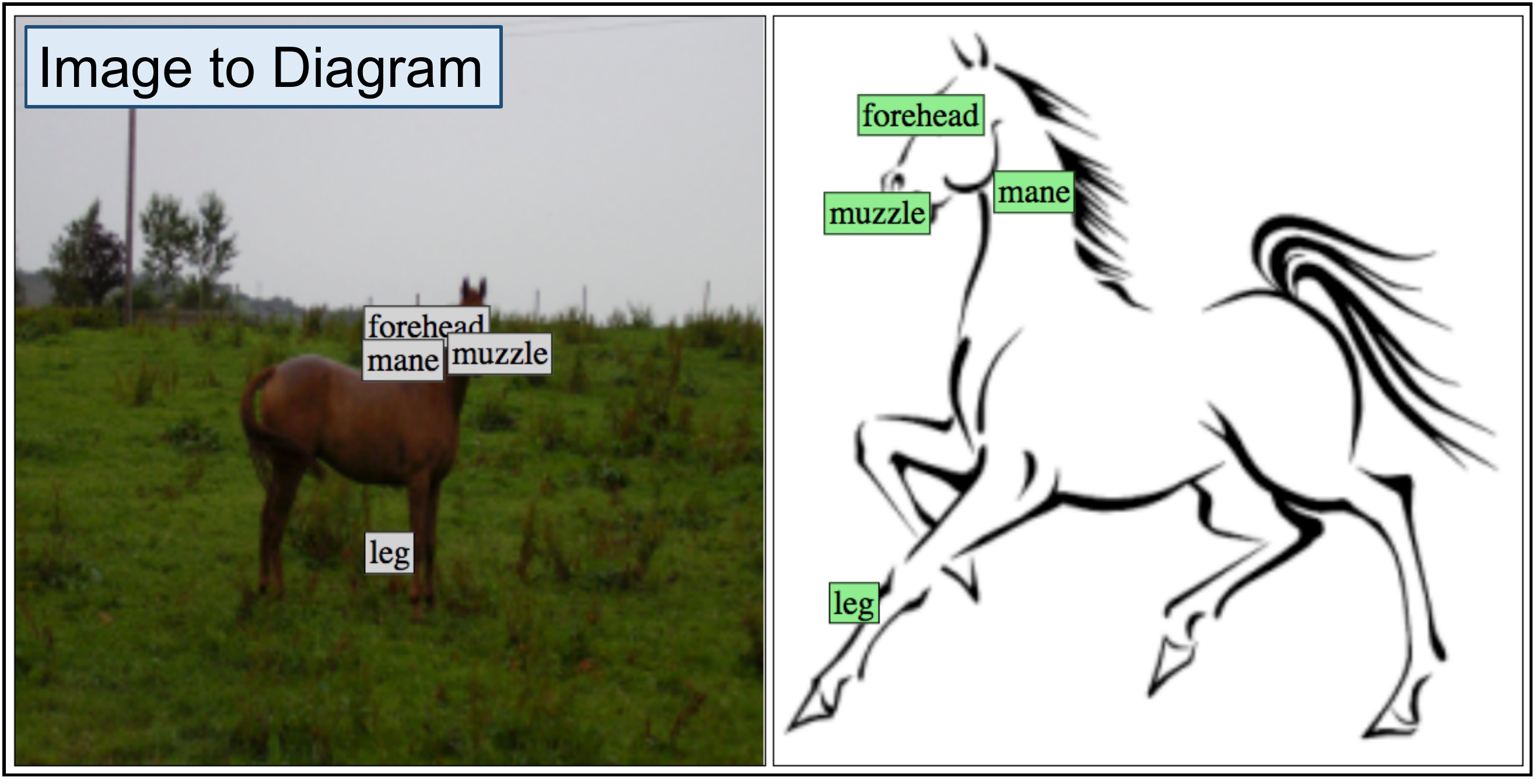}    
    \caption{Qualitative results from our SSMN model. In each pair of images, the labeled source images is on the left and the target is on the right. A green box indicates a correct match and a red box indicates an incorrect match.}
    \vspace{-1em}
    \label{fig:quali_result}
\end{figure*}

\begin{figure*}
    \centering
    \includegraphics[width=0.95\linewidth]{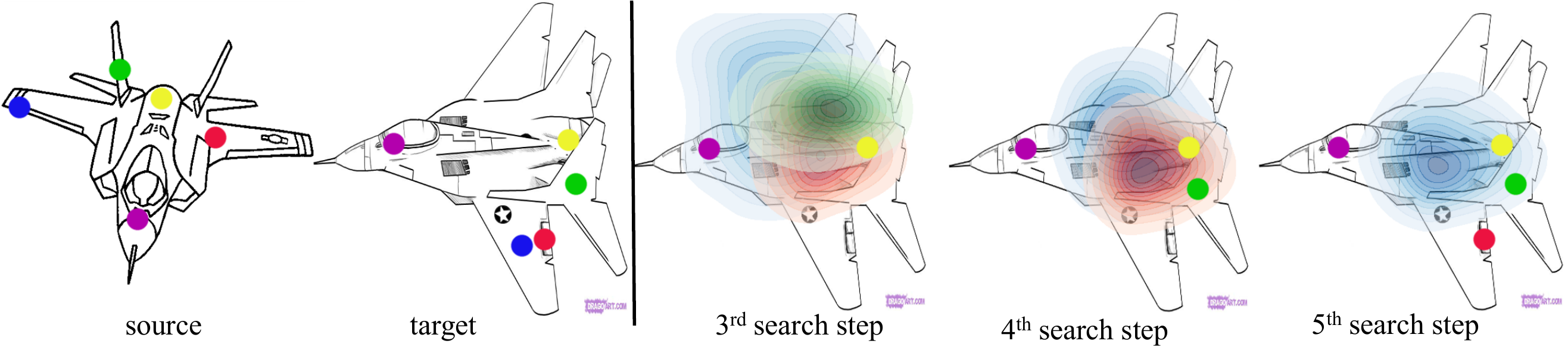}
    \caption{Visualization of expected part locations during the search by the structural consistency factor ($f_{sc})$.
    The example has both the varying pose and a non-trivial transformation of source to target part locations.
    The matched parts of each diagram are represented by the color-coded points.
    Each color-coded heatmap shows the score assigned by the structural factor to every location in the diagram for the unmatched part with the same color.
    For visual clarity, we display 5 of the 10 parts.
    }
    \vspace{-1em}
    \label{fig:structural_factor}
\end{figure*}

\noindent\textbf{Set-up.}
Training neural networks with global normalization typically requires pre-training with log-likelihood to obtain good results \cite{Andor2016,Wiseman2016SequencetoSequenceLA}. In training the SSMN, we pretrained it using the appearance matching network (AMN) surrogate objective (Sec~\ref{sec:training}), then fixed the weights of the convolutional layers while retraining the remainder of the network with 1 epoch of LaSO with a beam size parameter of 5 in most experiments unless mentioned (as described in Sec~\ref{sec:training}). We found that additional training epochs did not improve accuracy, presumably because pre-training brought the parameters close to a good solution. At test time, we ran the SSMN with a beam size of 100 in all experiments, except the ones that measured accuracy at different beam sizes. We chose this number based on experiments on the validation set which found that accuracy plateaued beyond 100.

The \emph{encoder network} uses two convolutional layers (64 filters of $3 \times 3$ and 96 filters of $3 \times 3$), each followed by $2 \times 2$ max pooling with stride of 2. This is followed by two fully connected layers, with 128 and 64 hidden units respectively. The \emph{context network} is a single layer bidirectional LSTM with 50 hidden units. We train the network using SGD with momentum and decay and $10^{-4}$ initial learning rate.

Note that the datasets we used in all the evaluations have no categories overlaps in train and test splits. It is a challenging set-up that matching the appearances of source and target that are never seen in training phase.

\begin{table}[t]
    \centering
    \begin{tabular}{lcc} 
        \toprule
        Methods $\backslash$ Dataset & \dataset\ & PPM \\
        \hline
        Random                      & 10.0\% & 10.0\%\\
        Nearest Neighbor (RGB)      & 29.4\% & 11.1\%\\
        Affine Transform            & 32.1\% & 26.9\%\\
        UCN~\cite{Choy2016UniversalCN}         
                                    & 38.9\% & 20.2\%\\
        Matching Network (MN)~\cite{vinyalsBLKW16}    
                                    & 41.3\% & 40.2\%\\
        $^\llcorner$ MN+Hungarian        
                                    & 45.6\% & 42.7\%\\ 
        \hline
        Appearance Matching Network+NN   
                                    & 35.7\% & 42.3\%\\ 
        SSMN-$f_{gc}$
                                    & 44.7\% & 40.6\%\\
        \textbf{SSMN (Ours)}
                                    & \textbf{58.1}\% & \textbf{46.6}\%\\
        \bottomrule
    \end{tabular}
    \caption{Accuracies of SSMN and other methods on both datasets.}
    \vspace{-1em}
    \label{tab:results_10part}
\end{table}

\vspace{.5em}
\noindent\textbf{Baselines.}
We compare SSMN to the following baselines.

\noindent \textbf{Nearest Neighbor (RGB)} computes matches using local appearance cues only by comparing raw image patches centered on the part's point using a euclidean metric.

\noindent \textbf{Affine Transform} baseline selects the matching of points that minimizes the error of a least-squares fit of the target part locations given the source part locations. This is not scalable to compute exactly, as it requires running a least-squares fit for every matching (3.7 million for 10 part matchings). We ran this approximately using beam search with width equal to 100.

\noindent \textbf{Matching Network (MN)}~\cite{vinyalsBLKW16} independently predicts a source point for each target point. This network runs the appearance matching network described in Section \ref{sec:appearance_similarity}, \ie. the encoder network with bidirectional LSTMs, to score each source given a target. The network is trained by feeding these scores into a $K$-way softmax then maximizing log-likelihood. A limitation of MN is that it does not enforce a 1:1 matching, hence may yield an invalid solution.

\noindent \textbf{MN + Hungarian} solves this problem by finding the maximum weighted matching given the matching network's scores. In contrast to the SSMN, this baseline uses the Hungarian algorithm as a post-processing step and is not aware of the matching constraint during training.


\noindent \textbf{Appearance Matching Network + NN} computes nearest neighbor matches using only appearance cues by the Appearance Matching network. Source and target points are fed into the encoders and matched using cosine similarity.

\noindent \textbf{Universal Correspondence Network (UCN)}~\cite{Choy2016UniversalCN} originates from the the semantic correspondence (SC) matching literature. 
Minimal post processing was required to adapt it to our task and compute an accuracy metric comparable to the SSMN and other baselines. 
Best results were obtained when fine tuning their pre-trained network on our datasets.

Table \ref{tab:results_10part} compares the accuracy of SSMN with the above baselines on the test sets for the DiPART and PPM datasets. The nearest neighbor baselines (both RGB and Appearance Matching Network) perform poorly, since they only use appearance cues with no contextual information and no matching constraint. 
The MN models outperforms all other baselines in both datasets. It clearly demonstrates that using sequential context, even in a set environment yields good results, consistent with the findings in~\cite{vinyalsBLKW16}.
Enforcing a 1:1 matching constraint via the Hungarian algorithm further improves this model. SSMN also outperforms UCN on both datasets. The SSMN outperforms other baselines because of its ability to model global consistency among the source and target sets. Training with global normalization is crucial for this improvement: if we train SSMN with local normalization, accuracy drops significantly (SSMN-$f_{gc}$).

\begin{figure}[h]
	\centering
    \includegraphics[width=0.95\linewidth]{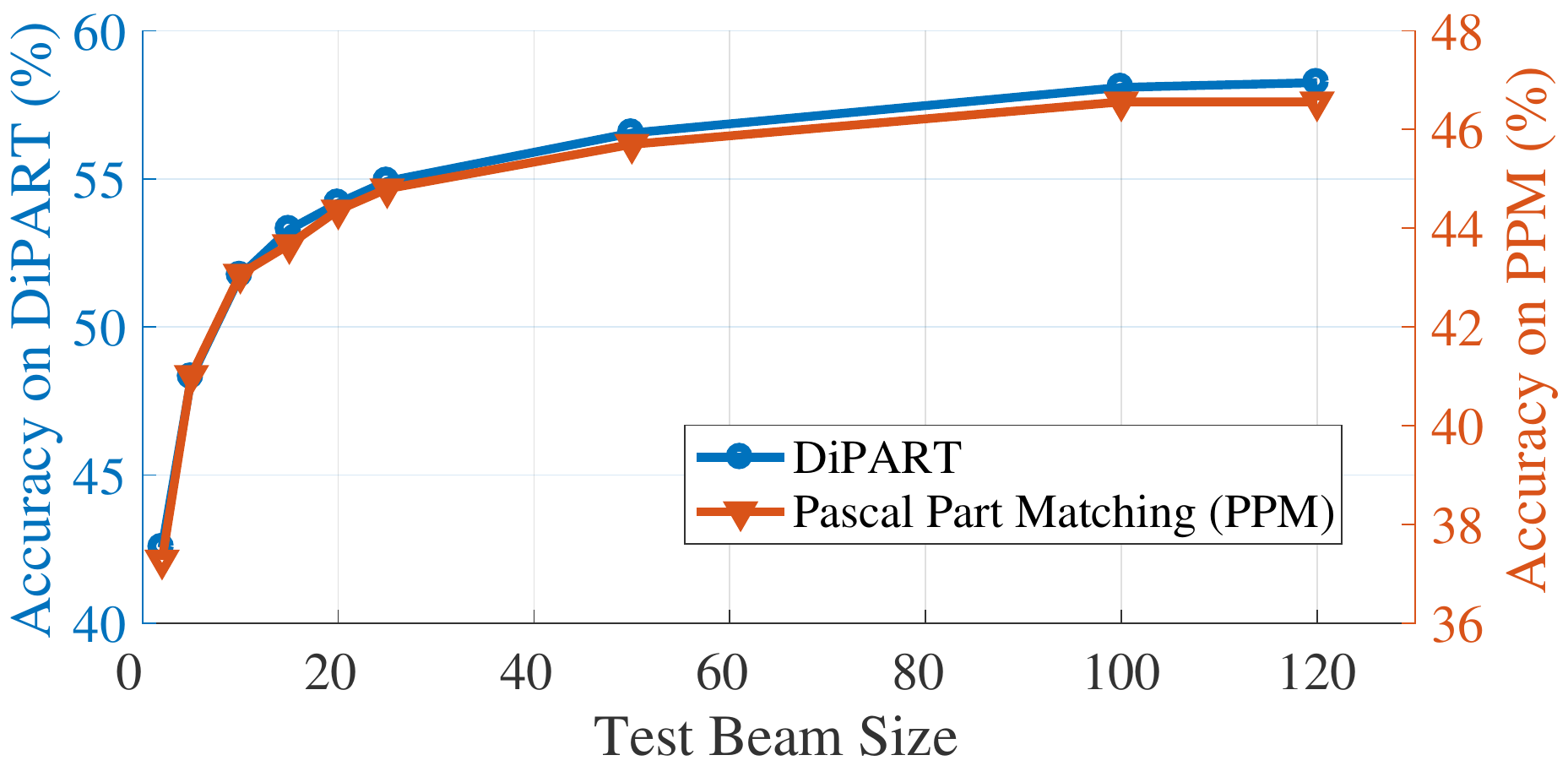}
    \caption{Accuracy as a function of inference beam size.}
    \vspace{-1em}
    \label{fig:beam_size}
\end{figure}

\vspace{.5em}
\noindent\textbf{Effect of Beam Size.}
Fig.~\ref{fig:beam_size} shows the test accuracies as a function of inference beam size. SSMN outperforms baselines even for beam sizes as low as 10, and saturates beyond 100. Note that even a beam size of 100 represents a tiny fraction ($0.0027\%$) of the search space of matchings ($10!$).

%

\vspace{.5em}
\noindent\textbf{Distance Transform (DT).}
We propose using DT images as inputs to our encoder networks, as opposed to the original diagrams. We compared the two approaches using just the appearance matching network, in order to isolate appearance cues from structural cues. Using DT images provides an accuracy of $38.4\%$ where as the original image produces $33.5\%$, a noticeable improvement. An interesting observation was that when we swept the space of filter sizes to find the best performing one for each configuration, the best filters for the original image were $15\times15$ as reported in~\cite{Yu2016SketchaNetAD} but the best filters for the DT image was $3\times3$, which is consistent with CNN architectures built for natural images.


\vspace{.5em}
\noindent\textbf{Does General Part Appearance Help?}
\label{sec:parts_help}
51\% of part names in the validation and 54\% in the test set appear in the training set of \dataset. Hence one might expect the part appearance factor ($f_p$) in SSMN to help significantly. An ablation study found that removing it caused very little drop in validation accuracy (within 0.1\%). This shows that, even though part names overlap significantly, part appearance cues do not always transfer between categories; \eg, a \emph{head} of an elephant and a giraffe look significantly different.

\vspace{.5em}
\noindent\textbf{Qualitative Analysis.}
Figure~\ref{fig:quali_result} shows qualitative examples and Figure~\ref{fig:structural_factor} visualizes SSMN's search procedure.

\vspace{.5em}
\noindent\textbf{Matching Variable Numbers of Parts.}
\dataset\ and PPM are setup to contain a fixed number of parts across images and a complete 1:1 matching between source and target sets. However, SSMN makes no such strict assumptions and can also be used in a relaxed setups. We modified \dataset\ to contain 9 parts in the source and 8 parts in the target image. 7 of these have a 1:1 matching and 1 part in each set has no correspondence in the counterpart. Table~\ref{tab:varypart} compares SSMN to the strongest baseline (MN+Hungarian). As 1:1 matching is not guaranteed, the Hungarian algorithm only marginally improves accuracy over MN while SSMN still provides large improvements.
\vspace{-.5em}
\begin{table}[h]
    \centering
    \begin{tabular}{cccc}
        \toprule
                        & MN        & MN+Hungarian  & SSMN       \\
        \hline
        Test Accuracy   & 31.1\%    & 31.5\%        & 38.2\% \\
        \bottomrule
    \end{tabular}
    \caption{Accuracies in DiPART with varying part setup.}
    \vspace{-0.5em}
    \label{tab:varypart}
\end{table}

\noindent\textbf{Cross Domain Matching.}
We evaluate cross domain matching on Cross-DiPART-PPM. This is the most challenging among the three setups. By using different \emph{encoder network} architecture for source and target, we demonstrate that SSMN is able to transfer labels across domains (images-to-diagrams) reasonably well. Since global geometric consistencies are preserved regardless of visual signatures, the SSMN outperforms the strongest baseline (Table~\ref{tab:crossdomain}). In this setup, the part classification term ($f_p$, Sec.~\ref{sec:part_appearance}) drops performance since the part classifiers do not generalize across domains. Thus, SSMN without $f_p$ (SSMN-$f_p$) provides further improvements.
\vspace{-0.5em}
\begin{table}[h]
    \centering
    \resizebox{8.2cm}{!}{
    \begin{tabular}{cccccc}
        \toprule
                        & Random & MN        & MN+Hungarian  & SSMN & SSMN-$f_p$\\
        \hline
        Test Accuracy   & 25\%   & 28.0\%    & 26.4\%        & 30.8\% & 33.1\% \\
        \bottomrule
    \end{tabular}
    }
    \caption{Cross domain accuracies (Cross-DiPART-PPM data)}
    \vspace{-0.5em}
    \label{tab:crossdomain}
\end{table}


\noindent \textit{More results can be found in the \href{https://s3-us-west-2.amazonaws.com/ai2-vision/one_shot_part_labeling/ssmn_supplementary.pdf}{supplementary material}}.
\vspace{-.5em}

\section{Conclusion}
\label{sec:conclusion}
We consider the challenging task of one-shot part labeling, or labeling object parts given a single example image from the category. We formulate this as set-to-set matching, and propose the Structured Set Matching Network (SSMN), a combined structured prediction and neural network model that leverages local appearance information and global consistency of the entire matching. SSMN outperforms strong baselines on three challenging setups: diagram-to-diagram, image-to-image and image-to-diagram.

\vspace{0.5em}
{\noindent \textbf{Acknowledgement.}
This work is in part supported by ONR N00014-13-1-0720, NSF IIS-1338054,  NSF-1652052, NRI-1637479, Allen Distinguished Investigator Award, and the Allen Institute for Artificial Intelligence.
JC would like to thank Christopher B. Choy (for the help in comparing with the UCN), Kai Han, Rafael S. de Rezende and Minsu Cho (for the discussion about SCNet) and Seunghoon Hong (for an initial discussion).
}

{\small
\bibliographystyle{ieee}
\bibliography{ssmn.bbl}
}

\end{document}